\documentclass[11pt,a4paper]{article}
\usepackage[hyperref]{emnlp2018}
\usepackage{times}
\usepackage{latexsym}

\usepackage{url}
\usepackage{amsmath,amssymb}
\usepackage{graphicx,graphics,subfig,epstopdf}
\usepackage{multirow}
\usepackage{floatrow}
\usepackage{makecell}
\usepackage{comment}

\def\x{{\mathbf x}}

\def\ben{\begin{equation}}
	\def\een{\end{equation}}

\aclfinalcopy 

\title{Large Margin Neural Language Model}

\author{
  Jiaji Huang$^1$\qquad\qquad
   Yi Li$^1$\qquad\qquad
   Wei Ping$^1$\qquad\qquad
   Liang Huang$^{1,2}$\thanks{\; Contributions were made while at Baidu Research.}\\
   $^1$ Baidu Research, Sunnyvale, CA, USA\\
   $^2$ School of EECS, Oregon State University, Corvallis, OR, USA\\
   {\small \tt \{huangjiaji, liyi17, pingwei01, lianghuang\}@baidu.com
   	}
   }

\date{}

\begin{document}
\maketitle
\begin{abstract}
We propose a large margin criterion for training neural language models. Conventionally, neural language models are trained by minimizing perplexity (PPL) on grammatical sentences.  However, we demonstrate that PPL may not be the best metric to optimize in some tasks, and further propose a large margin formulation. The proposed method aims to enlarge the \emph{margin} between the ``good" and ``bad" sentences in a task-specific sense. It is trained end-to-end and can be widely applied to tasks that involve re-scoring of generated text. Compared with minimum-PPL training, our method gains up to 1.1 WER reduction for speech recognition and 1.0 BLEU increase for machine translation.
\end{abstract}

\section{Introduction}
\label{sec:intro}
Language models~(LMs) estimate the likelihood of a symbol sequence $\{x^t\}_{t=0}^T$, based on the joint probability,
\begin{align}
\label{eq:LM}
p(x^0,\dots, x^T) = p(x^0) \prod_{t=1}^T p(x^t | x^0, \dots, x^{t-1}).
\end{align}
To measure the quality of an LM, a commonly adopted metric is perplexity (PPL), defined as
\[
\mbox{PPL}  \triangleq  \exp\left\{ -{1\over T}\sum_{t=0}^T\log p(x^t | x^0, \dots, x^{t-1})\right\},
\]
A good language model has a small PPL, being able to assign higher likelihoods to sentences that are more likely to appear.

LMs are widely applied in  automatic speech recognition (ASR)~\cite{Yu2014} and machine translation (MT)~\cite{Koehn2009}. Following~\citet{Koehn2009}, one may interpret the language model as prior knowledge on the text to be inferred, which provides information complementary to the ASR or MT system itself. In practice, there are several ways to incorporate the language model. The simplest way may be re-scoring an $n$-best list returned by the ASR or MT system~\cite{Mikolov2010, Sundermeyer2012}.
A slightly more sophisticated way is to jointly consider the ASR/MT and language model in a beam search decoder~\cite{DS2}.  Specifically, at each time step, the decoder appends every symbol in the vocabulary to each sequence in the current candidate set. For every hypothesis, a score is calculated as a linear combination of the log-likelihoods given by both the ASR/MT and language models. Then, only the top $K$ hypotheses with the highest scores are retained, as an updated candidate set.
More recently, \citet{Gulcehre2015} and \citet{Sriram2017} propose to predict the next symbol based on a fusion of the hidden states in the ASR/MT and language models. A gating mechanism is jointly trained to determine how much the language model should contribute.

The afore-discussed language models are generative in the sense that they merely model the joint distribution of a symbol sequence (Eq.~\eqref{eq:LM}). While the research community is mostly focused on pushing the limit of PPL~\citep[e.g.,][]{Jozefowicz2016}, very limited attention has been paid to the discrimination power of language models when they are applied to real tasks, such as ASR and MT~\cite{Li2008}. By contrast, discriminative language modeling aims at enhancing the performance in downstream applications. For example, existing works~\citep{Roark2004, Roark2007} often target at improving ASR accuracy. The key motivation underlying them is that the model should be able to discriminate between ``good"  and ``bad" sentences in a task-specific sense, instead of just modeling grammatical ones. The common methodology~\cite{Dikic2013} is to build a binary classifier upon hand-crafted features extracted from the sentences. However, it is not obvious how these methods can utilize large unannotated corpus, which is often easily available, and the hand-crafted features are also ad hoc and may result in suboptimal performance.

In this work, we study how to improve the discrimination ability of a recurrent network-based neural language model (RNNLM). The goal is to enlarge the difference between the log-likelihoods of ``good" and ``bad" sentences. In an contrast to the existing works~\citep{Roark2004, Roark2007}, our method does not rely on hand-crafted features, and is trained in end-to-end manner and able to take advantage of large external text corpus. In fact, it is a general training criterion that is transparent to the network architecture of the RNNLM, and can be applied to various text generation tasks, including ASR and MT. Experiments on state-of-art ASR and MT systems show its significant advantage over an LM trained by minimizing PPL.


\section{Background on RNNLM}
We first give some background knowledge on RNNLMs. The prototypical RNNLM~\cite{Mikolov2010} has one layer of recurrent cell and works as follows. Denote a sentence as $\mathbf x = [x^0, \dots, x^t, \dots]$, where the $x^t$'s are words. Let $\vec x^t$ be the embedding vector for $x^t$. The recurrent cell takes in the embedding and produces a hidden state $\vec h^t$ by
\[
	\vec h^t = \sigma(U \vec x^t + V \vec h^{t-1}),
\]
where $\sigma(z) = \frac{1}{1+e^{-z}}$ is sigmoid activation function. $\vec h^{t-1}$ is the hidden state at the last timestep. $U$ and $V$ are learnable parameters. The $\vec h^t$ is then passed into a multi-way classifier to produce a probability distribution over the vocabulary (for the next word),
\[
	\vec p = \mbox{softmax} (W \vec h^t + \vec b).
\]
The $W$ and $\vec b$ are also trainable parameters. The training objective is to maximize the log-likelihood of the next word, and the parameters are learned by back-propagation algorithm.

The vanilla recurrent cell can also be replaced by one or multiple layers of LSTM cells, which produces better results~\cite{Zaremba2014}. In a more general form, the RNNLM can be represented as a conditional probability, $p_\theta(x^t| x^0, \dots, x^{t-1})$, parameterized by $\theta$. In the prototypical case, $\theta=[U, V, W, \vec b]$. We could define the \emph{LM-score} of a sentence $\x$ as
\[\begin{aligned}
\mbox{LM-score}(\x) &\triangleq \log p_\theta(\x) \\
&= \sum_t \log p_\theta(x^t| x^0, \dots, x^{t-1}).
\end{aligned}\]
The RNNLM is trained by maximizing the average LM-score over all the $\x$'s in a corpus, or equivalently, minimizing the PPL on the corpus.

\section{Problem Formulation}
\label{sec:formulation}
We motivate and formulate a large margin training criterion in this section.
Suppose for every reference sentence $\x_i$, we have a collection of hypotheses $\x_{i,j}, j=1,\dots, K$, usually obtained as the top-$K$ candidates by a beam search decoder.

\subsection{A Motivating Example}
\label{sec:motivating_example}
An RNNLM trained by minimizing PPL cannot guarantee a higher score on the ``gold" reference than the inferior hypothesis, which is undesirable. One example is given in Tab.~\ref{tab:motivating_example}. The reference is taken from the text labels of dev93' set of Wall Street Journal (WSJ) dataset. The hypothesis is generated by a CTC-based~\cite{Graves2006} ASR system trained on WSJ training set. Words in red are mistakes made by the hypothesis. We then train an RNNLM on Common Crawl\footnote{\small\url{ http://web-language-models.s3-website-us-east-1.amazonaws.com/wmt16/deduped/en-new.xz}} copora by minimizing PPL. Training follows a typical setup~\cite{Jozefowicz2016} with a vocabulary of 400K the most frequent words. Any out-of-vocabulary word is replaced by an $\langle$UNK$\rangle$ token. The RNNLM is then employed to score the sentences. The LM-score of the erroneous hypothesis is higher than that of the reference. In fact, this is reasonable as ``a decade as concerns" seems to be a more common phrase.
In the training corpus, we find that ``a decade as concerns" appears once, but ``its defeat is confirmed" does not appear. Moreover, ``a decade as" appears 2,280 times, but ``its defeat is" appears only 24 times.
However, this is undesirable because if there is another hypothesis that happens to be the same as reference, which will not be ranked as the best candidate.
\begin{table}[t!]
\setlength{\tabcolsep}{2.5pt}
\centering
\begin{tabular}{c|c|c}
	\hline\hline
		& Sentence  & LM-score\\
	\hline
	\small{reference} & 
			\small{\thead{coniston declined to discuss its\\ plans for its defeat is confirmed\\ but indicated that it doesn't\\ plan to simply walk away}} & -116.52\\
	\hline
	\small{hypothesis} & 
			\small{\thead{coniston declined to discuss its\\ plans for \textcolor{red}{a decade as concerns}\\ but indicated that it doesn't\\ plan to simply walk away}} & -112.65\\
	\hline\hline
\end{tabular}
\caption{Reference and one hypothesis, scored by an RNNLM. Words in red are mistakes in the hypothesis. The RNNLM is trained on Common Crawl copora by minimizing PPL. We want the reference to be higher scored than the hypothesis, but it does not happen here.}
\label{tab:motivating_example}
\end{table}
\begin{figure}[t!]
	\centering
	\subfloat[]{\label{fig:naive_loss}\includegraphics[width=0.45\columnwidth]{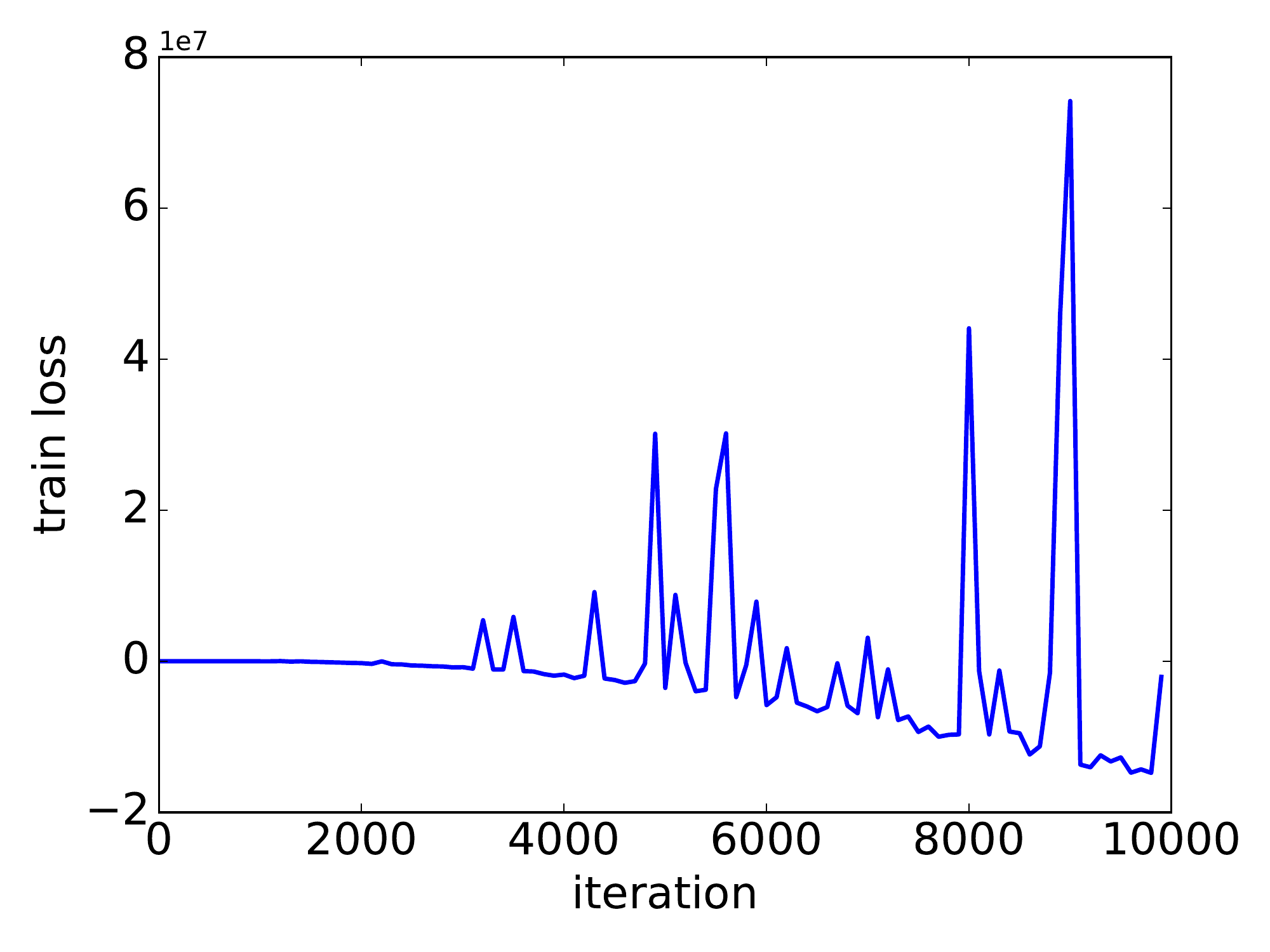}}\quad
	\subfloat[]{\label{fig:LMLM_loss}\includegraphics[width=0.45\columnwidth]{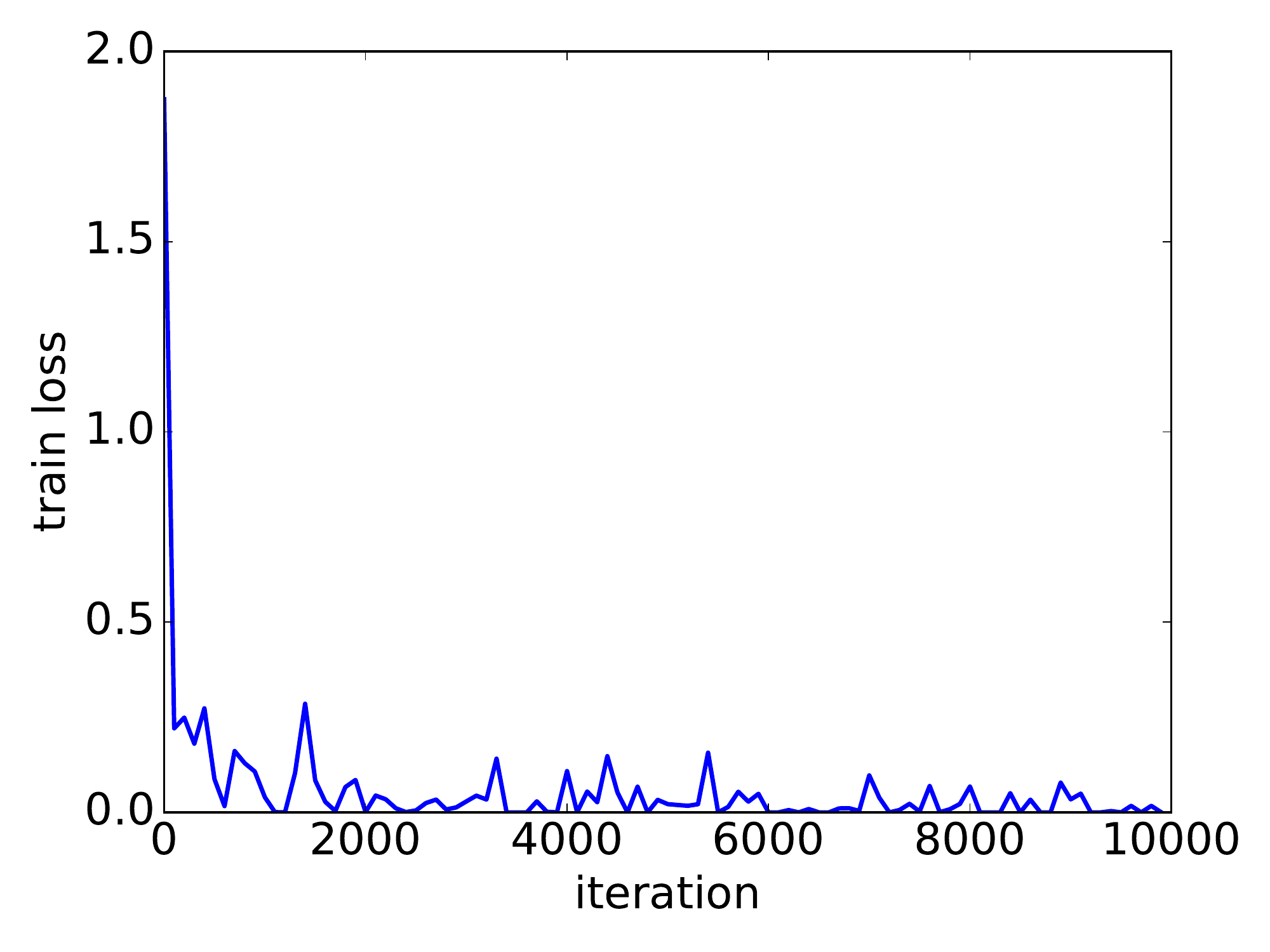}}
	\caption{Training losses of (a) straightforward formulation Eq.~\eqref{eq:naive}; and (b) large margin formulation Eq.~\eqref{eq:LMLM}}
	\label{fig:naive-vs-LMLM-loss}
\end{figure}
\begin{figure}[t!]
	\centering
	\subfloat[]{\label{fig:naive_margin}\includegraphics[width=0.45\columnwidth]{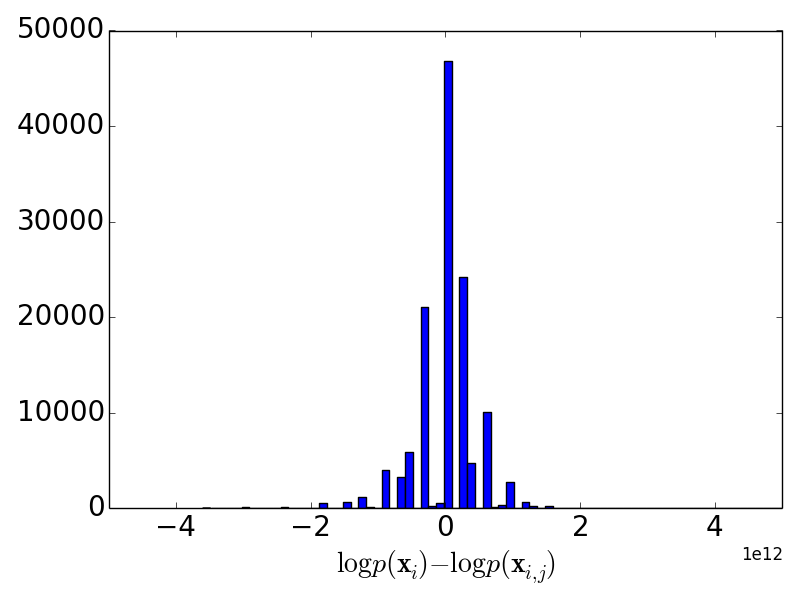}}\quad
	\subfloat[]{\label{fig:LMLM_margin}\includegraphics[width=0.45\columnwidth]{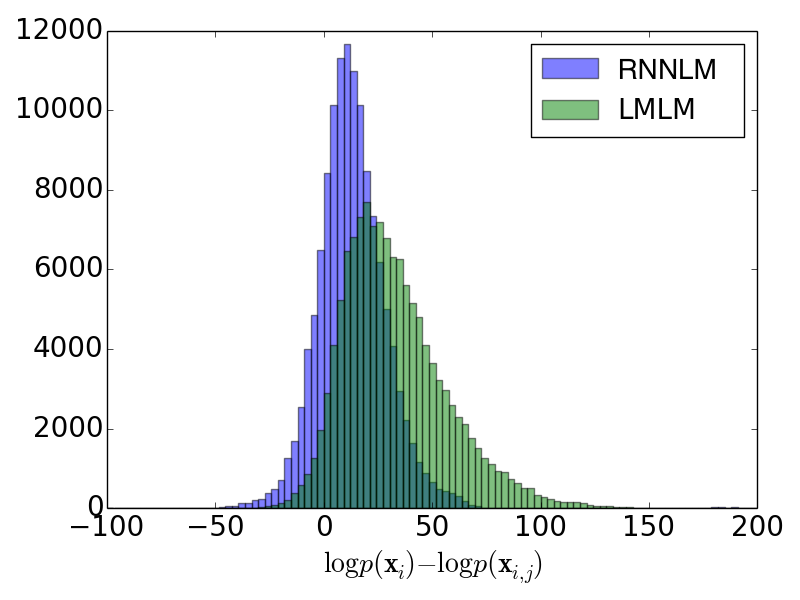}}
	\caption{Histogram of the margin $\log p(\x_i) - \log p(\x_{i, j})$. The more positive, the more the discrimination. (a) Straightforward formulation; (b) LMLM compared with RNNLM (a minimum-PPL LM trained on Common Crawl)}
	\label{fig:naive-vs-LMLM}
\end{figure}

It would be helpful if the LM can also learn from the imperfect hypotheses so that it can tell apart ``good" and ``bad" candidates. With this motivation, we train to assign larger LM-scores for the $\x_i$'s but smaller ones for the (imperfect) $\x_{i,j}$'s. A quantity of particular interest is $\log p(\x_i) - \log p(\x_{i,j})$, the \emph{margin}/difference between the LM-scores of the references and the (imperfect) hypotheses. The intuition is that the more positive the margin, the better the LM is at discrimination.

\subsection{Straightforward but Failed Formulation}
\label{sec:naive}
Without loss of generality, we assume that all the $\x_{i,j}$'s are imperfect and different from $\x_i$. A straightforward way is to adopt the following objective:
\ben
\min_\theta {1\over N}\sum_{i=1}^N\left( -\log p_\theta(\x_i) + \frac{1}{K} \sum_{j=1}^K\log p_\theta(\x_{i,j}) \right).
\label{eq:naive}
\een
Similar formulation is also seen in~\cite{Watanabe2015}, where they only utilize one beam candidate, \textit{i.e.}, $K=1$. Optimization can be carried out by mini-batch stochastic gradient descent~(SGD). Each iteration, SGD randomly samples a batch of $i$'s and $j$'s, computes stochastic gradient w.r.t. $\theta$, and takes an update step. However, a potential problem with this formulation is that the second term (corresponding to the inferior hypotheses) may dominate the optimization. Specifically, the training is almost always driven by the $\x_{i, j}$'s, but does not effectively enhance the discrimination. We illustrate this fact in the following experiment.

Using the ASR system in section~\ref{sec:motivating_example}, we extract 256 beam candidates for every training example in Wall Street Journal (WSJ) dataset. Warm started from the pre-trained RNNLM in section~\ref{sec:motivating_example}, we apply SGD to minimize the loss in Eq.~\eqref{eq:naive}, with a mini-batch size of 128.
The training loss is shown in Fig.~\ref{fig:naive_loss}. We observe that the learning dynamic is very unstable, and deceases to be negative. The unbound decreasing is due to the second term in Eq.~\eqref{eq:naive} being negative and dominating the training process. Next, we inspect $\log p_\theta(\x_i) - \log p_\theta(\x_{i, j})$, the \emph{margin} between the scores of a ground-truth and a candidate. In Fig.~\ref{fig:naive_margin}, we histogram the margins for all the $i, j$'s in a dev set. The distribution appears to be symmetric around zero, which indicates poor discrimination ability. Given these facts, we conclude that the straightforward formulation in Eq.~\eqref{eq:naive} is not effective.

\subsection{Large Margin Formulation}
To effectively utilize all the imperfect beam candidates, we propose the following objective,
\ben
\min_\theta \sum_{i=1}^N\sum_{j=1}^B \max \big\{ 0, \tau-(\log p_\theta(\x_i) - \log p_\theta(\x_{i, j})) \big\},
\label{eq:LMLM}
\een
where $\log p_\theta(\x_i) - \log_\theta(\x_{i, j})$ is the margin between the scores of a ground-truth $\x_i$ and a candidate $\x_{i, j}$.
The hinge loss on the margin encourages the log-likelihood of the ground-truth to be at least $\tau$ larger than that of the imperfect hypothesis. We call an LM trained by the above formulation as Large Margin Language Model (LMLM).

We repeat the same experiment in section~\ref{sec:naive}, but change the objective function to Eq.~\eqref{eq:LMLM} and set $\tau=1$. Fig.~\ref{fig:LMLM_loss} shows the training loss, which steadily decreases and approaches zero rapidly. 
Compared with the learning curve of naive formulation (Fig.~\ref{fig:naive_loss}), the large margin based training is much more stable. In Fig.~\ref{fig:LMLM_margin}, we also examine the histogram of $\log p_\theta(\x_i) - \log p_\theta(\x_{i, j})$, where $p_\theta(\cdot)$ is now the LM learned by LMLM. Compared with the histogram by the conventional RNNLM, LMLM significantly moves the distribution to the positive side, indicating more discrimination.

\subsection{Ranking Loss Type Formulation}
In most cases, all beam candidates are imperfect. It may be beneficial to exploit the information that some candidates are relatively better than the others. We consider ranking them according to some metrics w.r.t.~the ground-truth sentences. For ASR, the metric is WER, and for MT, the metric is BLEU score. We define $\x_{i,0}\triangleq \x_i$ and assume that the candidates $\{\x_{i,j}\}_{j=1}^K$ are sorted such that
\[
\mbox{WER}(\x_i, \x_{i, j-1}) < \mbox{WER}(\x_i, \x_{i, j})
\]
for ASR, and
\[
\mbox{BLEU}(\x_i, \x_{i, j-1}) > \mbox{BLEU}(\x_i, \x_{i, j})
\]
for MT. In other words, $\x_{i, j-1}$ has better quality than $\x_{i,j}$.

We then enforce the ``better" sentences to have a score at least $\tau$ larger than those ``worse" ones. This  leads to the following formulation,
\ben
\begin{aligned}
\min_\theta \sum_{i=1}^N & \sum_{j=0}^{B-1}\sum_{k=j+1}^B \max \big\{ 0, \\
	& \tau-(\log p_\theta(\x_{i,j}) - \log_\theta(\x_{i, k})), \big\}.
\end{aligned}
\label{eq:rankLMLM}
\een
Compared with LMLM formulation Eq.~\eqref{eq:LMLM}, the above introduces more comparisons among the candidates, and hence more computational cost during training. We call this formulation ranking-loss-based LMLM (rLMLM).

To summarize this section, we have proposed LMLM and rLMLM that aim at discriminating between hypotheses in a task-specific (e.g., WER or BLEU)  sense, instead of minimizing PPL.

\section{Experiments on ASR}
We apply the LMs trained under different criteria to rescore the beams in various ASR systems. In particular, we are interested in knowing which of the two training mechanisms is better: minimizing PPL (e.g., the RNNLM in Section~\ref{sec:motivating_example}), or fitting to the WER metric by the proposed methods.

Adapting an RNNLM to a specific domain has been of interest, especially to the speech community~\cite{Park2010, Chen2015, Ma2017}. We adopt~\citet{Ma2017} that fine-tune the softmax layer of RNNLM by minimizing the PPL on the text labels of training set. According to~\citet{Ma2017}, the reason not to fine-tune all the layers is due to the  limited text labels in the target domain. Indeed, we also observe overfitting if adapting all layers, but adapting only the softmax layer effectively decreases the PPL on the text labels of dev sets. We refer to this fine-tuning as \emph{RNNLM-adapted} in the following sections.

To make a fair comparison with the adapted model, we also use the RNNLM as an initialization for our LMLM and rLMLM. In total, there are four language models for rescoring the beams. RNNLM and its adapted version that aim at reducing PPL; and the two proposed methods, LMLM and rLMLM that try to fit to WER.

\subsection{WSJ Dataset}
\label{sec:wsj}
The WSJ corpora consists of about 80 hours 
of read speech with texts drawn from a machine-readable
corpus of Wall Street Journal news. We use the standard
configuration of train si284 dataset for training, dev93 for
development and eval92 for testing.

Our ASR model has one convolution layer, followed by 5 bidirectional RNNs and one fully connected layer, with a CTC loss on top. The text labels of the training set are used to train a 4-gram language model, which is employed in the ASR decoder. The beam search decoder has a beam width of 2000. Before beam rescoring, this ASR system achieves a WER of 12.16 on dev93 set and 7.69 on eval92 set. To put this into perspective, we list some previous state-of-the-art system in Tab.~\ref{tab:WSJ ASRs}. Compared with them, our baseline is already very competitive.

\subsubsection{WERs and PPLs}
The out-of-vocabulary rate of WSJ text is only 0.28\%, making the RNNLM reasonable to use.
We apply the RNNLM, RNNLM-adapted~\cite{Ma2017}, LMLM and rLMLM to rescore the beams on dev and test set. The final score assigned to a beam is a weighted sum of the ASR and language model scores. The weight is found by minimizing the WER on the dev set.

Tab.~\ref{tab:WSJ_rescore} reports the WERs on dev93 and eval92 sets. All methods reduce the WER over the baseline without rescoring. However, LMLM and rLMLM are notably better than the other two methods. Moreover, although RNNLM and RNNLM-adapted achieve smaller PPLs on the text labels, the advantage does not transfer to WER. 
\begin{table}[h!]
	\setlength{\tabcolsep}{2pt}
	\begin{tabular}{c|c|c}
		\hline\hline
		\multirow{ 2}{*}{ASR Models} & \multicolumn{2}{c}{WER}\\
		\cline{2-3}
		& dev93 & eval92\\
		\hline
		EESEN~\cite{Miao2015} & N/A & 7.34\\
		Attention~\cite{Bahdanau2016} & N/A & 9.30\\
		Gram-CTC~\cite{hairong2017}& N/A & 6.75\\
		5-layer Bidi-RNNs (baseline) & 12.16 & 7.69\\
		\hline\hline
	\end{tabular}
	\caption{Published WERs on WSJ dev93 and eval92 set}
	\label{tab:WSJ ASRs}
\end{table}
\begin{table}[h!]
	\setlength{\tabcolsep}{2.5pt}
	\begin{tabular}{c|r|c|c|c}
		\hline\hline
		rescoring & \multicolumn{2}{c|}{WER} & \multicolumn{2}{c}{PPL}\\
		\cline{2-5}
		language model & \multicolumn{1}{c|}{dev93} & \multicolumn{1}{c|}{eval92} & \multicolumn{1}{c|}{dev93} & \multicolumn{1}{c}{eval92}\\
		\hline
		baseline & \multirow{2}{*}{12.16} & \multirow{2}{*}{7.69} & \multirow{2}{*}{N/A} & \multirow{2}{*}{N/A} \\
		(no rescore)  & & & &\\
		\hline
		RNNLM & 10.71 & 6.59 & 207.43 & 205.00\\
		RNNLM-adapted & 10.11 & 6.34 & 159.50 & 157.85\\
		LMLM & {\bf 9.44} & {\it 5.56} & 575.83 & 563.69\\
		rLMLM & {\it 9.63} & {\bf 5.48} & 345.60 & 348.32\\
		\hline\hline
	\end{tabular}
	\caption{Rescore 2000-best list of WSJ dev93 and eval92 set. Digits in bold are the best and italics are the runner-ups. Lower PPL does not correspond to lower WER.}
	\label{tab:WSJ_rescore}
\end{table}
\begin{figure}[h!]
	\includegraphics[width=0.8\columnwidth]{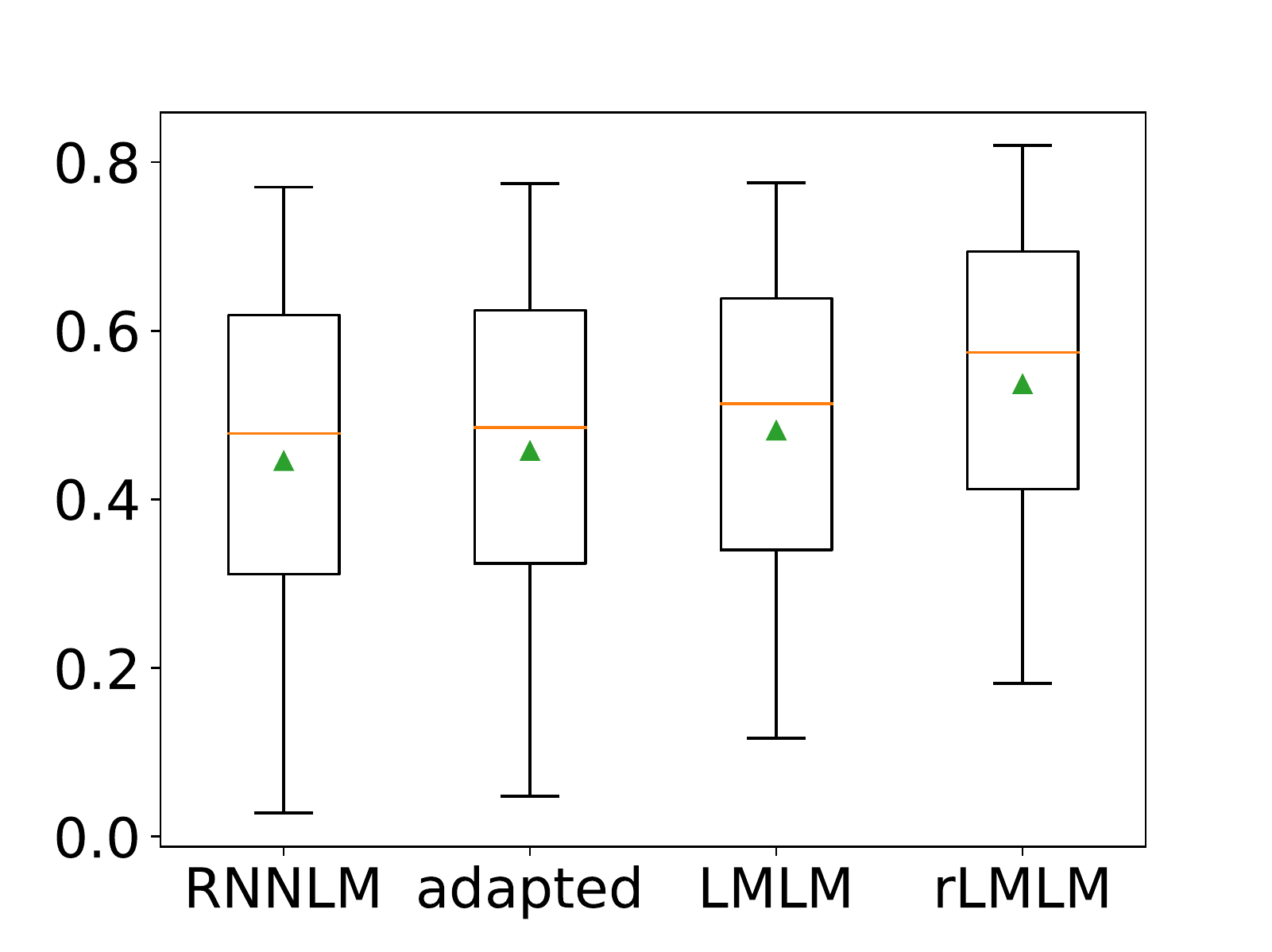}
	\caption{Correlation coefficients between word level accuracy ($1-\mbox{WER}/100$) and LM-scores by the different LMs, higher is better. Red horizontal lines are medians. Green dots are means. Whiskers are 5\% and 95\% quantiles. Lower and upper box boundaries are 25\% and 75\% quantiles.}
	\label{fig:corr_wer_metric}
\end{figure}

\subsubsection{Correlation between scores and WERs}
\label{sec:corr}
To better understand the proposed methods, we calculate the correlation coefficients between the hypotheses' WERs and their scores (by different language models). In specific, for every utterance in the test set, we have a set of beam candidates, their word level accuracies (100-WER) and scores given by an LM, from which a Pearson correlation coefficient can be calculated. We calculate the coefficients for all the utterances in the test set, and boxplot these coefficients in Fig.~\ref{fig:corr_wer_metric}. The correlation coefficients by LMLM and rLMLM tend to be higher than RNNLM and RNNLM-adapted. This indicates that LMLM and rLMLM are more aligned with the goal of reducing WER.
\begin{table*}[t]
	\centering
	\setlength{\tabcolsep}{2pt}
	\begin{tabular}{l|c|c|c|c|c}
		\hline\hline
		\multicolumn{1}{c|}{\multirow{2}{*}{references and hypotheses}} & \multirow{2}{*}{\thead{reference or\\ ranked 1st by}} & \multicolumn{4}{c}{\emph{LM-score}}\\
		\cline{3-6}
		& & \small{RNNLM} & \thead{RNNLM\\-adapted} & \small{LMLM} & \small{rLMLM} \\
		\hline
		\thead{for such group rate coverage employers can charge the \\
			former workers and their families the average cost of providing\\
			the health benefits plus a two percent administrative fee} & \thead{reference,\\ LMLM} & -144.74 & -142.90 & -162.61 & -165.93 \\
		\thead{for such group rate coverage employers can charge the\\
			 former workers and their families the average cost of providing \\
			 the health benefits plus \textcolor{red}{\textvisiblespace} two percent administrative \textcolor{red}{\textvisiblespace}} & \thead{RNNLM,\\ RNNLM-adapted} & -144.53 & -142.85 & -172.66 & -168.10 \\
		\thead{for such group rate coverage employers can charge the \\
			former workers and their families the average cost of providing \\
			\textcolor{red}{their} health benefits plus a two percent administrative fee} & \thead{rLMLM} & -146.72 & -145.67 &-163.73 & -162.92\\
		\hline
		\thead{we'd like to see something that leads to real democracy\\ says jaime bonilla vice secretary general} & reference & -105.8 & -106.28 & -122.93 & -108.42\\
		\thead{we'd like to see something that leads \textcolor{red}{the} real democracy\\ says \textcolor{red}{jm bonier} vice secretary general} & \thead{RNNLM,\\ RNNLM-adapted}& -103.13 & -102.84& -141.94 & -118.38\\
		\thead{we'd like to see something that leads to real democracy\\ says \textcolor{red}{jim bone} vice secretary general} & \small{LMLM, rLMLM} & -104.52 & -105.37 & -125.40 & -104.73 \\
		\hline
		\small{the big shoe is going to drop when we see the trade number} & reference & -64.28 & -61.04 & -80.63 & -82.91\\
		\small{the big \textcolor{red}{she was} going to drop \textcolor{red}{in to} see the trade \textcolor{red}{numbers}} &\thead{RNNLM, \\ RNNLM-adapted}&-64.32 & -61.68 & -94.26 & -86.89\\
		\small{the big shoe is going to drop \textcolor{red}{in} \textcolor{red}{\textvisiblespace} see the trade number} & \small{LMLM, rLMLM}& -67.53 & -64.50 & -84.09 & -84.65\\
		\hline\hline
	\end{tabular}
	\caption{Some ``gold" references and best hypotheses (after rescoring by different language models) for eval92 set. In red are errors or missing word (denoted as `\textcolor{red}{\textvisiblespace}').}
	\label{tab:WSj_beams}
\end{table*}

\subsubsection{Case Study}
Tab.~\ref{tab:WSj_beams} posts some examples from the test set. The first column lists the ground-truth labels, and their corresponding best candidates as re-ranked by the four LMs (see notes in the second column). Words in red are mistakes made by the candidate sentences. Scores of these sentences are listed in the last four columns. We have the following observations:
\begin{enumerate}
	\item LMLM and rLMLM give worse scores on the ground-truth labels than RNNLM and RNNLM-adapted, which explains their higher PPL in Tab.~\ref{tab:WSJ_rescore}.
	\item In the first example, RNNLM and RNNLM-adapted assign higher scores to a shorter sentence. This is reasonable (though not necessarily desirable) as LM-score is a summation of log-probabilities, each of which is negative. In contrast, LMLM and rLMLM are able to assign higher scores to longer and better candidates.
	\item In the other two examples, LMLM and rLMLM seem to favor more sensible sentences, though they are not more grammatical than those picked by RNNLM and RNNLM-adapted. We conjecture that since LMLM and rLMLM utilize beam candidates in their training, they capture and compensate for some weakness in the ASR, which is not achieved by RNNLM and RNNLM-adapted.
\end{enumerate}

\subsection{10K Speech Dataset}
We further validate our methods on a larger noisy dataset collected by~\citet{hairong2017}. The dataset has about 10K hours of spontaneous speech. The utterances are corrupted by background noise, and a large portion of them are accented. Therefore it is much more challenging than WSJ. We adopt the same training-dev-test split as in \citet{hairong2017}. In specific, there are 5.4M utterances for training, 2,066 for development and 2,054 for testing.

\begin{table}[h!]
	\centering
	\setlength{\tabcolsep}{2.5pt}
	\begin{tabular}{c|c|c|r|r}
		\hline\hline
		rescoring & \multicolumn{2}{c|}{WER} & \multicolumn{2}{c}{PPL}\\
		\cline{2-5}
		language model & dev & test & \multicolumn{1}{c|}{dev} & \multicolumn{1}{c}{test}\\
		\hline
		baseline & \multirow{2}{*}{19.17} & \multirow{2}{*}{20.90} &\multicolumn{1}{c|}{\multirow{2}{*}{N/A}} & \multicolumn{1}{c}{\multirow{2}{*}{N/A}} \\
		(no rescore) & & & &\\
		\hline
		RNNLM & 18.38 & 20.07 & 264.21 & 252.85\\
		RNNLM-adapted & 18.29 & 20.03 & 236.74 & 226.22\\ 
		LMLM & {\it 18.17} & {\it 19.62} & 2250.79 & 2095.39\\
		rLMLM & {\bf 17.98} & {\bf 19.49} & 1225.04 & 1152.63\\
		\hline\hline
	\end{tabular}
	\caption{Rescore 2000-best list of our internal dev and test set. Digits in bold are the best and italics are the runner-ups.}
	\label{tab:ds2.1}
\end{table}

The ASR we build has the same architecture as in~\citet{hairong2017}, except that its decoder integrates an in-domain 5-gram language model. This system achieves a WER of 19.17 on dev set, better than the reported 19.77 baseline in~\citet{hairong2017}. Based on the ASR, we repeat the same experiments in section~\ref{sec:wsj}. Tab.~\ref{tab:ds2.1} reports WERs and PPLs on dev and test sets. Both LMLM and rLMLM outperform the other methods in WER, although their PPLs are higher. This trend is similar to that in Tab.~\ref{tab:WSJ_rescore}. 

\section{Experiments on NMT}
In this section, we experiment the large-margin criterion trained LM with a competitive Chinese-to-English NMT system. The NMT model is trained from 2M parallel sentence pairs. Following \citet{Shen2016}, we use NIST 06 newswire portion (616 sentences) for development and NIST 08 newswire portion (691 sentences) for testing. We use OpenNMT-py\footnote{\small\url{https: //github.com/OpenNMT/OpenNMT- py}} package with the default configuration to train the model: batch size is 64; word embedding size is 500; dropout rate is 0.3; target vocabulary size is 50K; number of epochs is 20, after which a minimum dev perplexity of 7.72 is achieved.

\subsection{BLEUs and PPLs}
We use a beam size of 10 for decoding, and report case-insensitive 4-reference BLEU-4 scores (by calling ``multi\_bleu.perl"\footnote{\small\url{ https://github.com/OpenNMT/OpenNMT-py/blob/master/tools/multi-bleu.perl}}). 
The NMT model achieves 35.18 BLEU score on dev set and 31.52 on test set (see table~\ref{tab:ch-en}). To put this into perspective, \citet{Shen2016} trains their models on 2.56M pairs of sentences and reports a dev BLEU score of 32.7 (via MOSES) or 30.7 (via RNNsearch, beam size of 10). So our NMT model is already very competitive.

To construct the training data for LMLM and rLMLM, 10 beam candidates are extracted for every sentence in the training set. We then follow the same experimental steps outlined in section~\ref{sec:wsj}, except that the ASR score is now changed to NMT score. In addition, we also find that normalizing the LM score by sentence length can improve the re-scoring performance substantially. Tab.~\ref{tab:ch-en} compares the BLEU score after re-ranking by the different LMs. LMLM and rLMLM both improve upon the baseline significantly, and outperform RNNLM and RNNLM-adapted by a notable margin. 
We also observe that the PPLs of LMLM and rLMLM are much larger than those of RNNLM and RNNLM-adapted, suggesting that the PPL metric may be very poorly correlated with BLEU.

Interestingly, RNNLM-adapted does not show any gain in BLEU score over RNNLM. To understand this, we recall that NMT is trained by minimizing PPL on target text. Its decoder is implicitly an RNNLM on target language. We conjecture that adapting an LM to the target domain can only duplicate the functionality of the NMT decoder, which does not bring any additional benefit.
\begin{table}[h!] 
	\centering
	\setlength{\tabcolsep}{2.5pt}
	\begin{tabular}{c|c|c|r|r}
		\hline\hline
		rescoring & \multicolumn{2}{c|}{BLEU} & \multicolumn{2}{c}{PPL}\\
		\cline{2-5}
		language model & dev06 & test08 & \multicolumn{1}{c|}{dev06} & \multicolumn{1}{c}{test08}\\
		\hline
		baseline & \multirow{2}{*}{35.18} 
		& \multirow{2}{*}{31.52 } 
		& \multicolumn{1}{c|}{\multirow{2}{*}{N/A}} & \multicolumn{1}{c}{\multirow{2}{*}{N/A}} \\
		(no rescore)  & & & & \\
		\hline
		RNNLM & 36.17  & 32.17  & 129.91 & 137.25 \\  
		RNNLM-adapted & 36.17  & 31.97  & 78.20 & 89.27 \\ 
		LMLM & {\it 37.79}  & {\it 33.11}  & 7.75e5  &  3.73e6 \\  
		rLMLM & {\bf 37.82}  & {\bf 33.13}  & 2.68e5 & 1.12e6 \\ 
		\hline\hline
	\end{tabular}
	\caption{Rescore 10-best list for dev (nist 06) and test (nist 08) set. Digits in bold are the best and italics are the runner-ups.}
	\label{tab:ch-en}
\end{table}
\begin{figure}
	\includegraphics[width=0.8\columnwidth]{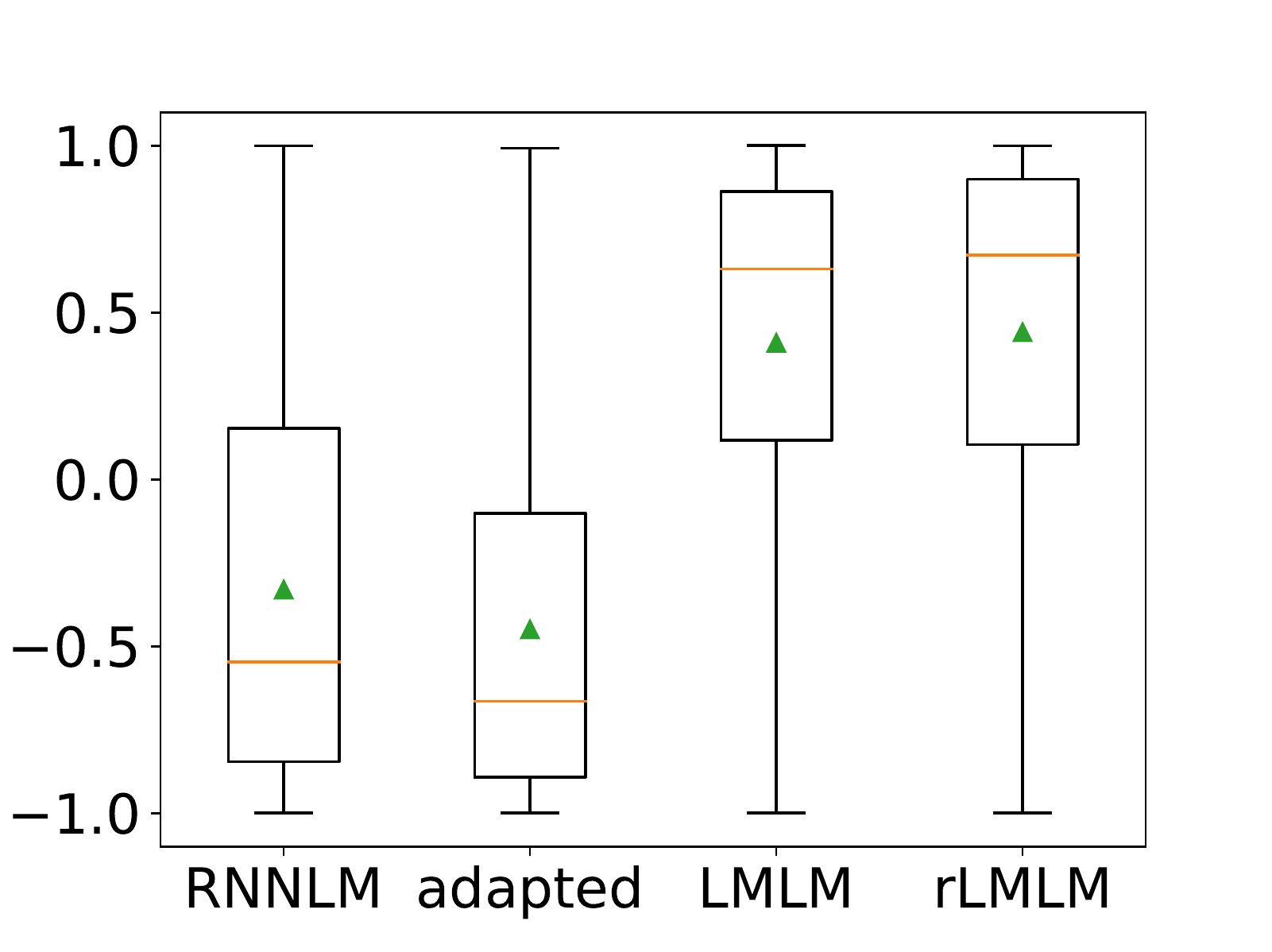}
	\caption{Correlation coefficients between BLEUs and LM-scores by the different LMs, higher is better. Red horizontal lines are medians. Green dots are means. Whiskers are 5\% and 95\% quantiles. Lower and upper box boundaries are 25\% and 75\% quantiles.}
	\label{fig:corr_nmt}
\end{figure}

\subsection{Correlation between scores and BLEUs}
We measure the correlation between the LM scores and BLUEs. The calculation is done on dev06 set in the same way as Section~\ref{sec:corr}, but now we change the WERs to BLEUs. The boxplot of the correlation coefficients are shown in Fig.~\ref{fig:corr_nmt}. Compared with the boxplot in Fig.~\ref{fig:corr_wer_metric}, now the correlation coefficients by all LMs are more dispersed. Sometimes, they even take negative values. The mean correlation by LMLM and rLMLM, however, is considerably higher than those by RNNLM and RNNLM-adapted.

\section{Related Work}
``\emph{Language modeling is an art of determining the probability of a sequence of words}"~\cite{Goodman2001}. In the past decades, there has been a trend of increasing the context that an LM can condition on. N-gram models~\cite{Chen1999} assume that each symbol depends on the previous $N-1$ symbols. Feed forward neural network based LMs \cite{Bengio2003} are not count based but they inherit the restrictive assumption. To model longer-term dependencies, RNNLMs \cite{Mikolov2010} are proposed.  RNNLMs often achieve smaller PPLs than the N-gram counterparts~\cite{Sundermeyer2012, Zaremba2014, Jozefowicz2016}. This paper focuses on RNNLM-type architectures.

While these works all adopt PPL as the metric to optimize, sometimes one may optimize a task-specific objective. For example, \citet{Kuo2002, Roark2007} and \citet{Dikic2013} propose discriminative LMs to improve speech recognition. The common methodology therein is to fit  a probabilistic model, e.g., conditional random field~\cite{Roark2004}, to the space of text candidates, and maximize the probability at the desired candidate. The problem is often solved by perceptron algorithm. However, these methods all rely on ad-hoc choice of features, e.g., counts of $n$-grams where $n$ varies in a small range (e.g.,1 to 3).  Moreover, it is also not clear how these methods would take advantage of an existing language model (trained on large unsupervised corpus). Nevertheless, the same methodology can be extended to RNNLMs, thus avoiding the aforementioned limitations. For example, \citet{Auli2014} train an RNNLM by favoring sentences with high BLEU scores and integrate it into a phrase-based MT decoder.

If we cast the problem of picking the best text sequence as a ranking problem, the aforementioned works can be considered as ``pointwise" learning-to-rank approaches~\cite{Cossock2008}. In contrast, the proposed method is a ``pairwise" approach~\cite{Liu2009}, as it learns a neural language model by comparison between pairs of sentences. Earlier works in this fashion may date back to \cite{Collins2005}, which improves a semantic parser. Learning ``by pairwise comparison" is also seen in several MT literatures. For example, \citet{Hopkins2011} propose to train a phrase-based MT system by minimizing a pairwise ranking loss. \citet{Wiseman2016} optimize the beam search process in a Neural Machine Translation (NMT) system. They enforce the score of a reference to be higher than that of its decoded k-th candidate by at least a unit margin.

Rather than optimizing the MT system itself, this work proposes a general method of training recurrent neural language models, which can benefit various text generation tasks, including speech recognition and machine translation.

\section{Conclusions}
We have proposed a large margin criterion for training recurrent neural language models. Rather than minimizing PPL, the proposed criterion is based on comparison between pairs of sentences. We have formulated two algorithms that implement the training criterion. One compares between references and imperfect hypotheses (LMLM), the other compares between all pairs of hypotheses (rLMLM). We applied the language models trained by these two algorithms to speech recognition and machine translation. Both of them demonstrate superior performance over their minimum-PPL counterparts. However, the performance gain from LMLM to rLMLM is small, although rLMLM is built on more pairwise comparisons and requires more training efforts. The efficiency with respect to the number of pairs is a future research topic.

\bibliographystyle{acl_natbib_nourl}
\bibliography{LMLM}

\end{document}